\documentclass[sigconf]{acmart}
\usepackage{siunitx}      
\usepackage{multirow}      
\usepackage{bm}           
\usepackage{xspace}       
\usepackage{booktabs} 
\usepackage{amsmath}  
\usepackage{array}    
\usepackage{calc}     
\usepackage{makecell}
\usepackage{placeins} 
\usepackage{pifont}

\renewcommand\footnotetextcopyrightpermission[1]{}
\AtBeginDocument{%
  }

\begin{document}

\title{MCOD: The First Challenging Benchmark for Multispectral Camouflaged Object Detection}

\author{%
Yang Li$^{1}$ \quad Tingfa Xu$^{1*}$ \quad Shuyan Bai$^{1}$ \quad Peifu Liu$^{1}$   \quad Jianan Li$^{1*}$\\
$^1$Beijing Institute of Technology\\
}

\begin{abstract}
Camouflaged Object Detection (COD) aims to identify objects that blend seamlessly into natural scenes. Although RGB-based methods have advanced, their performance remains limited under challenging conditions. Multispectral imagery, providing rich spectral information, offers a promising alternative for enhanced foreground-background discrimination. However, existing COD benchmark datasets are exclusively RGB-based, lacking essential support for multispectral approaches, which has impeded progress in this area. To address this gap, we introduce \textbf{MCOD}, the first challenging benchmark dataset specifically designed for multispectral camouflaged object detection. MCOD features three key advantages: \textbf{(i) Comprehensive challenge attributes:} It captures real-world difficulties such as small object sizes and extreme lighting conditions commonly encountered in COD tasks. \textbf{(ii) Diverse real-world scenarios:} The dataset spans a wide range of natural environments to better reflect practical applications. \textbf{(iii) High-quality pixel-level annotations:} Each image is manually annotated with precise object masks and corresponding challenge attribute labels. We benchmark eleven representative COD methods on MCOD, observing a consistent performance drop due to increased task difficulty. Notably, integrating multispectral modalities substantially alleviates this degradation, highlighting the value of spectral information in enhancing detection robustness. We anticipate MCOD will provide a strong foundation for future research in multispectral camouflaged object detection. The dataset is publicly accessible at \url{https://github.com/yl2900260-bit/MCOD}.

\end{abstract}



\maketitle

\section{Introduction}
Camouflaged Object Detection (COD) has recently emerged as a growing research area focused on identifying objects that are visually integrated into their surrounding environments. This task holds broad applicability and strategic importance in domains such as military reconnaissance, medical diagnostics, and agricultural pest monitoring~\cite{r4,r5,r6,r7}.

Driven by advances in deep learning, RGB-based COD methods have achieved notable progress~\cite{r8,r9,r10}. However, COD remains inherently challenging. A primary difficulty arises when objects share highly similar color or texture characteristics with the background. Additionally, factors such as extreme illumination, cluttered scenes, or small object sizes frequently lead to mispredictions by existing methods~\cite{r11,r12}. To address these limitations, researchers have explored techniques including multi-source information fusion and multi-task learning~\cite{r13,r14,r15}. While effective to some extent, these methods often require high computational resources and access to large-scale, diverse datasets.

\begin{figure}
  \includegraphics[width=1\linewidth]{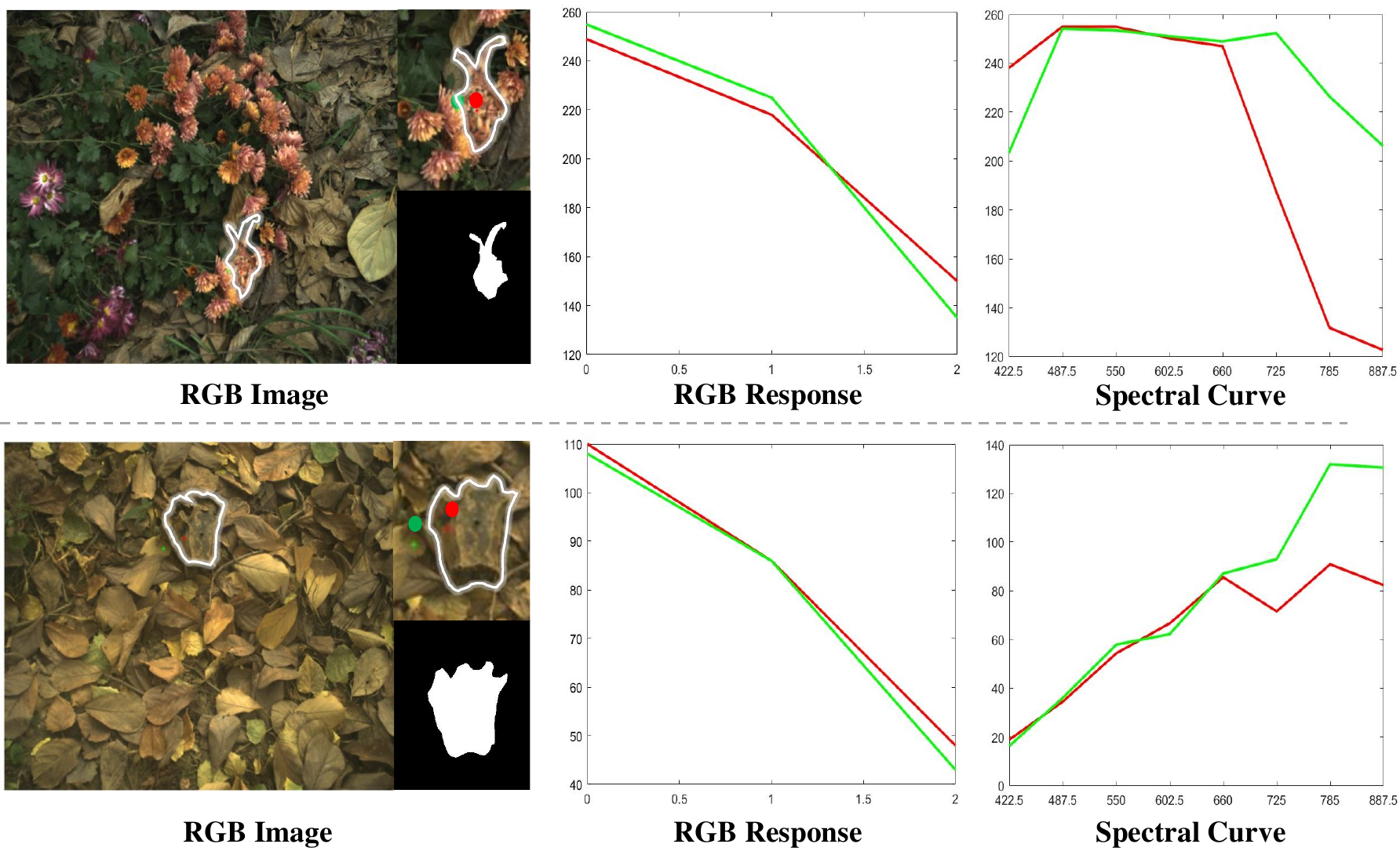}
  \vspace{-2mm}
  \caption{RGB response curves and spectral signatures of selected representative camouflaged scenes captured by a multispectral imaging camera. }
  \label{fig:MSI}
  \vspace{-10pt}
\end{figure}

Multispectral data, which captures both spatial and spectral characteristics of a scene, offers enhanced foreground-background discrimination in complex environments. As illustrated in Figure~\ref{fig:MSI}, objects that appear indistinguishable in RGB images can exhibit distinct spectral signatures in multispectral bands. Nevertheless, current deep learning-based COD approaches largely remain restricted to RGB modalities, and studies leveraging other modalities are still scarce. To the best of our knowledge, there is currently no publicly available benchmark dataset for camouflaged object detection based on multispectral imagery~\cite{r16}.

To overcome the limitations of existing datasets—such as single-modality inputs and insufficiently diverse challenge attributes—and to advance research in Multispectral Camouflaged Object Detection (MS-COD), we introduce \textbf{MCOD}, the first benchmark dataset specifically designed for camouflaged object detection using multispectral modalities. The key characteristics of MCOD are as follows:
\begin{itemize}
    \item \textbf{Comprehensive Challenge Attributes:} MCOD incorporates eight challenging attributes to facilitate the development of more robust detection models. Compared to existing RGB-based COD benchmarks, MCOD features significantly smaller objects, with an average object size of just 0.429\% of the image area. It also introduces distinctive challenges, including extreme illumination conditions and object displacement. These diverse attributes are designed to more accurately reflect real-world complexities.
    
    \item \textbf{Diverse Real-world Scenarios:} MCOD samples are collected from diverse natural scenes, such as urban parks, leaf-covered surfaces, paved roads, grassy fields, and landscaped gardens, thereby enhancing the dataset's generalizability and practical relevance.
    
    \item \textbf{Fine-grained Annotations:} High-quality annotations are critical for COD tasks. All labels in MCOD are manually annotated and refined through multiple rounds of human review to ensure pixel-level accuracy. The annotation process required over 1,800 labor-hours to generate detailed ground-truth masks and attribute tags for each image.
\end{itemize}

To evaluate the performance of existing COD methods on MCOD and to establish baselines for future research, we conduct a comprehensive assessment of eleven representative models. We further analyze the effect of incorporating multispectral modalities into the COD framework. Experimental results demonstrate that state-of-the-art models exhibit significant performance degradation on the more challenging MCOD dataset. Notably, the inclusion of multispectral information effectively alleviates this decline.

By introducing MCOD, we aim to foster a new research direction focused on learning more discriminative camouflage features from multispectral data, advancing both theoretical understanding and practical applications of COD. In summary, our contributions are: (i) We propose MCOD, the first benchmark dataset for COD with multispectral modalities; (ii) MCOD provides high spatial and spectral resolution and diverse challenging attributes to support robust models; and (iii) we benchmark eleven representative methods, demonstrating the value of multispectral imagery in enhancing COD performance and offering a strong foundation for future work.

\section{Related Works}
\noindent
\textbf{Camouflaged Object Detection Datasets. }Several influential open-source benchmark datasets, such as CHAMELEON~\cite{r2}, CAMO-COCO~\cite{r17}, COD10K~\cite{r18}, and NC4K~\cite{r19}, have greatly promoted the development of camouflaged object detection based on RGB images. CHAMELEON is a small-scale, non-peer-reviewed dataset consisting of 76 camouflaged images collected from the internet via keyword searches. CAMO-COCO includes 2,500 images, among which the CAMO subset provides 1,250 images with seven challenging attributes designed to increase detection difficulty. NC4K is currently the largest publicly available image-level COD test set, containing 4,121 camouflaged object images also sourced from the internet. COD10K is the most comprehensive image-level COD dataset to date, containing 5,066 camouflaged images, with category annotations for all camouflaged instances. However, due to the limited spectral resolution of RGB images, these datasets struggle to distinguish objects from highly similar backgrounds.

\noindent
\textbf{Camouflaged Object Detection Methods.} With the advent of deep learning, COD based on RGB images has made significant progress~\cite{r16, r21}. Methods such as SINet~\cite{r11}, C2FNet~\cite{r22}, and PENet~\cite{r23} leverage attention mechanisms to enhance feature focusing capabilities, substantially improving detection performance. Nevertheless, attention mechanisms alone are insufficient to capture fine-grained object structural features. To address this, methods like PopNet~\cite{r24} and BGNet~\cite{r14} utilize multi-source information fusion to incorporate depth or boundary cues for improved contour reconstruction. In addition, prompt-based learning approaches such as VSCode~\cite{r13} and GenSAM~\cite{r25} adaptively guide and enhance COD models using textual or visual prompts to address modality-specific challenges. Despite their effectiveness, these methods remain confined to the RGB modality, overlooking the spectral features that can significantly enhance the foreground-background separability by capturing intrinsic object properties.

\noindent
\textbf{Multispectral Camouflaged Object Detection}\textbf{.} Large-scale datasets such as MSRS~\cite{r26} and KAIST~\cite{r27} have demonstrated the effectiveness of multispectral information in object detection and significantly advanced the field. Unfortunately, COD research remains largely constrained to RGB-based benchmarks. To the best of our knowledge, no publicly available benchmark dataset currently exists for camouflaged object detection using multispectral imagery. From a data collection standpoint, RGB images can be conveniently gathered from search engines and open databases, making the construction of RGB-based datasets relatively easy. In contrast, multispectral data acquisition requires professional imaging equipment, which makes building multispectral COD datasets significantly more challenging. This limitation has inevitably hindered the progress of multispectral camouflaged object detection.

\section{MCOD Dataset}

\begin{table*}[h] 
    \centering
    \caption{Comparison of the MCOD dataset with other COD datasets. \textit{Avg-Pix.}: Average number of pixels; \textit{Attr.}: Number of included attributes; \textit{Attr.-Label}: Whether attribute labels are provided; \textit{Camo-Img.}: Images containing camouflage objects.}
    \label{tab:dataset}
    \vspace{-1mm}
    \begin{tabular}{c|cccccccc}
        \toprule
        \textbf{Dataset} & \textbf{Year} & \textbf{Modality} & \textbf{Channels} & \textbf{Avg-Pix.} & \textbf{Resolution} & \textbf{Attr.}  &\textbf{Attr.-Label}& \textbf{Camo-Img.} \\
        \midrule
        CHAMELEON\cite{r2} & 2017 & RGB & 3 & 902474 & \makecell{\hfill 450×300~\textasciitilde{}2304×34 \hfill} & -  &\ding{55}& 76 \\
        CAMO-COCO\cite{r17} & 2019 & RGB & 3 & 552440 & \makecell{\hfill 154×156~\textasciitilde{}7360×4912 \hfill} & 7  &\ding{55}& 1250 \\
        NC4K\cite{r19} & 2021 & RGB & 3 & 391150 & \makecell{\hfill 354×268~\textasciitilde{}1280×960 \hfill} & -  &\ding{55}& 4121 \\
        COD10K\cite{r18} & 2022 & RGB & 3 & 724442 & \makecell{\hfill 300×199~\textasciitilde{}2976×3968 \hfill} & 7  &\ding{51}& 5066 \\
        \midrule
        \textbf{MCOD} (Ours) & 2025 & MSI & \textbf{8} & \textbf{980400} & 1140×860 & \textbf{8}  &\ding{51}& 1527 \\
        \bottomrule
    \end{tabular}
\end{table*}

\begin{figure*}
  \centering
  \includegraphics[width=1\linewidth]{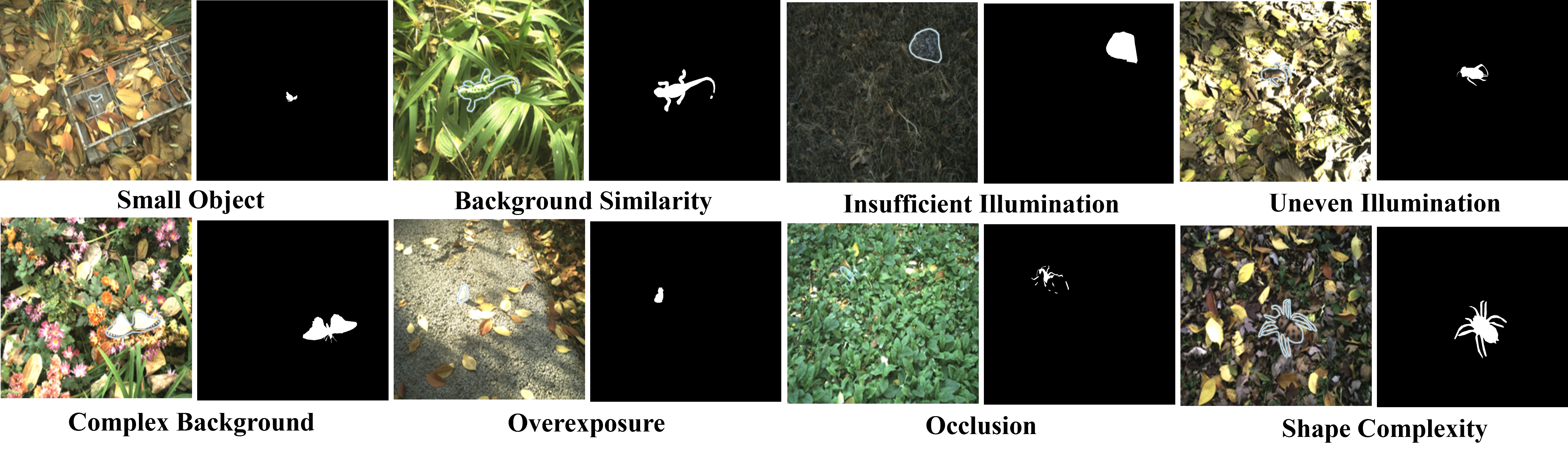}
  \vspace{-4mm}
  \caption{Illustration of the eight challenging attributes in the MCOD dataset.}
  \label{fig:attributes_overview}
  \vspace{-5pt}
\end{figure*}

\subsection{Construction Principle}

We propose \textbf{MCOD} as a novel benchmark dataset designed to advance the field of COD by providing high spatial and spectral resolution multispectral images enriched with diverse challenging attributes and fine-grained annotations. To this end, we adhere to the following principles in constructing MCOD:

\noindent
\textbf{A Richer Set of Challenging Attributes.} A diverse set of challenging attributes is essential for improving the robustness and generalization of COD models. We expect MCOD to include more challenging attributes, thereby enabling the training of more powerful COD models on our dataset.

\noindent
\textbf{A Broader Range of Real-world Scenarios.} A key motivation of MCOD is to promote more generalizable camouflaged object detection through extensive real-world scenarios. To this end, we aim for the new benchmark dataset to cover a wider range of real-world scenes, including natural backgrounds commonly found in practical environments.

\noindent
\textbf{Strict Image Quality Control.} To ensure data quality, we conducted a rigorous screening of the collected images. Multispectral images that were misaligned, blurry, incorrectly calibrated, or exhibited poor camouflage were excluded from MCOD.

\noindent
\textbf{High-quality Image Annotations.} In the context of COD tasks, the level of detail in image annotations directly affects the training and evaluation performance of models. To ensure the high quality of MCOD, all images are provided with pixel-level ground truth labels and manually annotated challenge attributes.

\subsection{Dataset Overview}

MCOD is the first benchmark dataset for camouflaged object detection based on multispectral imagery, designed to facilitate progress in the emerging field of multispectral camouflaged object detection. Table~\ref{tab:dataset} provides a comprehensive comparison of MCOD with several widely-used benchmark datasets across multiple dimensions. Compared to other commonly used benchmarks in the camouflaged object detection field, MCOD demonstrates clear advantages in data modality, average image resolution, and the diversity of challenging attributes. Specifically, MCOD captures the inherent complexity of real-world camouflage scenarios by incorporating eight distinct challenging attributes, surpassing datasets such as COD10K and CAMO-COCO, which each contain seven. As noted in~\cite{r37}, high-resolution imagery offers richer object boundary details, enhancing model training and ultimately improving test-time performance. MCOD features an average of \textbf{980400} pixels per image—the highest among existing COD benchmarks. Most notably, MCOD is the first dataset to introduce 8-channel multispectral imagery to the COD task, expanding research from traditional RGB modalities to multispectral ones and enabling a systematic exploration of multispectral camouflaged object detection.

\subsection{Dataset Construction}

\noindent
\textbf{Dataset Collection. }The primary principle guiding our data collection process is to provide a comprehensive set of radiometrically calibrated and spatially registered multispectral and RGB images that cover a wide range of real-world scenarios and diverse challenging attributes. To this end, we collected over 4,200 multispectral images across various natural environments. To ensure data quality, we conducted a rigorous screening process on the collected images. First, the raw images were calibrated, and the 2nd, 3rd, and 5th spectral bands from the calibrated multispectral images were used to reconstruct false-color composites. We then discarded images that were misaligned, blurred, improperly calibrated, or exhibited poor camouflage quality. After this meticulous filtering process, a total of 1,527 high-quality multispectral images were retained.

\noindent
\textbf{Dataset Annotation.} We adhered to annotation protocols commonly used in existing COD datasets, providing both pixel-level ground-truth masks and challenge attribute labels for all images. All annotations were generated using MATLAB’s ImageLabeler toolbox. Compared to conventional RGB-based datasets, the annotation process for MCOD presented additional challenges. First, MCOD includes smaller camouflaged objects, making precise pixel-level labeling more difficult. Second, it features a wide range of challenging scenarios, such as complex object shapes, overexposed regions, and high background similarity, which further complicate object localization and contour delineation. 

To address these difficulties, we leveraged 8-channel multispectral images to enhance object visibility under complex conditions and to facilitate accurate annotation. All images in the MCOD dataset were manually labeled, with all annotations verified by domain experts. Over 1,800 labor hours were dedicated to the meticulous annotation and review process to ensure high precision and consistency.

\noindent
\textbf{Dataset Splitting.} The MCOD dataset comprises 1,527 multispectral images, divided into a training set of 1,027 images and a testing set of 500 images. To maintain consistency, we ensured that the training and testing sets share similar data distributions. A thorough manual inspection was conducted to prevent potential data leakage. Notably, allocating 500 images (approximately one-third of the dataset) to the test set ensures sufficient diversity and challenge, enabling a more comprehensive evaluation of camouflaged object detection models and a robust assessment of their generalization across diverse scenarios.

\subsection{Dataset Statistics}
\begin{figure}
  \includegraphics[width=1\linewidth]{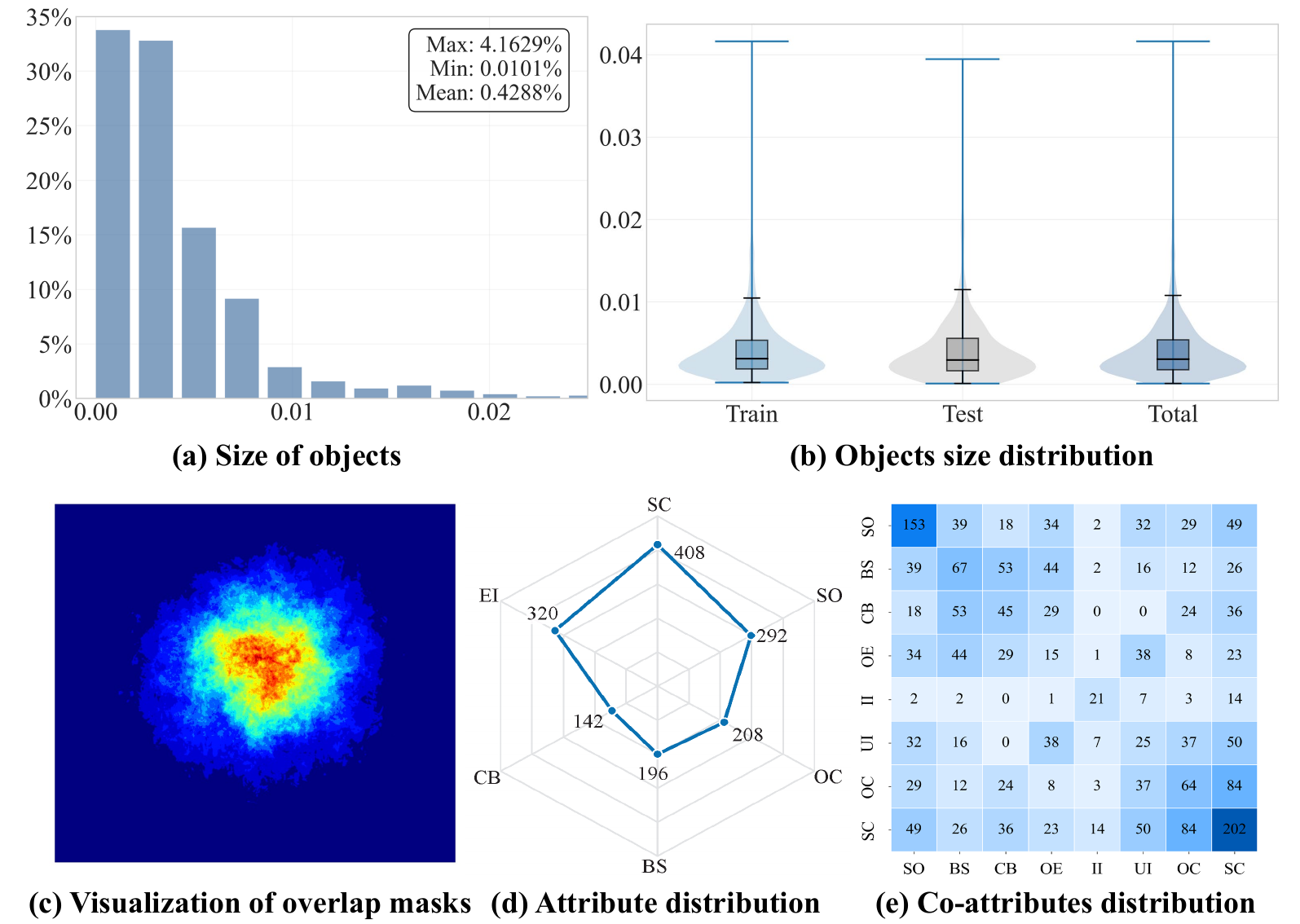}
  \caption{Statistical analysis of the MCOD dataset. }
  \label{fig:statistics}
 \vspace{-6pt}
\end{figure}

Figure~\ref{fig:statistics} provides a detailed presentation of our statistical analysis of MCOD from multiple perspectives.

\begin{table}
    \centering
\caption{Definitions of the Challenge Attributes in MCOD}
\label{tab:attribute-definitions}
    \begin{tabular}{cl}\toprule
         Attr.& Description\\\midrule
         \textbf{SC}& \textit{Shape Complex}: Objects have complex boundaries.\\
         \textbf{SO}& \textit{Small Object}: Object-to-image area ratio < 0.001. \\
         \textbf{OC}& \textit{Object Occlusion}: Object is partially occluded.\\
         \textbf{BS}& \textit{Background Similarity}: Background similar to object.\\
         \textbf{UI}& \textit{Uneven Illumination}: Unevenly illuminated object.\\
         \textbf{CB}& \textit{Complex Background}: Object in complex background.\\
         \textbf{OE}& \textit{Overexposure}: Object excessively exposed to light.\\
         \textbf{II}& 	\textit{Insufficient Illumination}: Object lacks adequate lighting.\\ \bottomrule
    \end{tabular}
    \vspace{-4mm}
\end{table}

\noindent
\textbf{Camouflaged Object Size.} 
The size of camouflaged objects has a substantial impact on detection performance, with smaller objects typically being more difficult to detect. We define object size as the ratio of camouflaged object pixels to the total number of image pixels. As illustrated in Figure~\ref{fig:statistics}(a), the MCOD dataset contains a high proportion of images with extremely small objects. The object sizes range from 0.01\% to 4.16\%, with an average of 0.429\%—significantly smaller than the 8.94\% reported in COD10K~\cite{r18}.

\noindent
\textbf{Object Size Distribution in Subsets.} To ensure similar object size distributions across the training and testing sets, we manually adjusted the data partitioning. As illustrated in Figure~\ref{fig:statistics}(b), both subsets exhibit nearly identical size distributions, maintaining statistical consistency with the overall dataset. This careful design supports fair and reliable model evaluation. Figure~\ref{fig:statistics}(c) visualizes the spatial distribution of object centers, where brighter areas indicate higher occurrence probabilities.

\noindent
\textbf{Attribute Distribution.} To support in-depth analysis of multispectral camouflaged object detection, we define eight challenge attributes in MCOD, with detailed descriptions provided in Table~\ref{tab:attribute-definitions}. For ease of statistical analysis, the attributes II (Insufficient Illumination), UI (Uneven Illumination), and OE (Overexposure) are collectively categorized under the EI (Extreme Illumination) attribute for quantitative evaluation. Figure~\ref{fig:statistics}(d) visualizes the frequency distribution of these attributes, while Figure~\ref{fig:statistics}(e) illustrates the co-occurrence of attributes, revealing their interrelationships. According to the results shown in the figure, the most prevalent challenge attributes in MCOD are complex shape, extreme illumination, and small object, collectively accounting for 65.1\% of the entire dataset. The inclusion of diverse and challenging attributes enables more comprehensive and accurate evaluation of existing COD models.

\section{Experiments}
In this section, we benchmark eleven state-of-the-art camouflaged object detection methods on the newly proposed MCOD dataset. All experiments were conducted on an NVIDIA GEFORCE RTX 3090 24GB GPU, with multispectral images as inputs. To enable these models to operate on multispectral inputs, we adapted the input layer by expanding the original 3-channel RGB input to 8 channels. The pretrained weights of the first layer were extended via 3D convolution to initialize the 8-channel weights.

We employ four widely used evaluation metrics to assess the performance of camouflaged object detection models: Mean Absolute Error (MAE or \(M\))~\cite{r28}, F-measure (\(F_\beta\))~\cite{r29}, Structure-measure (\(S_\alpha\))~\cite{r30}, and Enhanced Alignment Measure (\(E_\xi\))~\cite{r31}.

\begin{table}[!t]
    \centering
\caption{Benchmarking results of 11 camouflage object detection models on the MCOD dataset. The best performance for each evaluation metric is highlighted in bold. }
\label{tab:benchmark}
\vspace{-2mm}
\begin{tabular}{c|c|cccc} 
    \toprule
    \textbf{Method} & \textbf{Venue} & \boldmath$E_{\xi}\uparrow$ & \boldmath$S_{\alpha}\uparrow$ & \boldmath$F_{\beta}\uparrow$ & \boldmath$M\downarrow$ \\
    \midrule
    SINet\cite{r11}& CVPR 2020 & 0.758 & 0.616 & 0.369 & 0.006 \\
    LSR\cite{r19}& CVPR 2021 & 0.830 & 0.625 & 0.373 & 0.005 \\
    CODCEF\cite{r34} & Sensors 2021 & 0.763 & 0.677 & 0.444 & 0.004 \\
    C2FNet\cite{r22}& IJCAI 2021 & 0.726 & 0.721 & 0.403 & 0.010 \\
    C2FNet-V2\cite{r33} & TCSVT 2022 & 0.913 & 0.810 & 0.654 & 0.008 \\
    SINet-V2\cite{r18} & TPAMI 2022 & 0.849 & 0.728 & 0.492 & 0.004 \\
    ASBI\cite{r38}& CVIU 2023& 0.684& 0.675& 0.370&0.014\\
    FIRNet\cite{r39}& TVC 2024& 0.882& 0.738& 0.537&0.004\\
    PRNet\cite{r36} & TCSVT 2024 & \textbf{0.926} & 0.826 & \textbf{0.698} & \textbf{0.002} \\
    IdeNet\cite{r35} & TIP 2024 & 0.846 & 0.808 & 0.588 & 0.004 \\
    PCNet\cite{r32} & arXiv 2024 & 0.633 & \textbf{0.855} & 0.386 & 0.003 \\
    \bottomrule
\end{tabular}

\end{table}

\begin{figure*}
  \includegraphics[width=1.\linewidth]{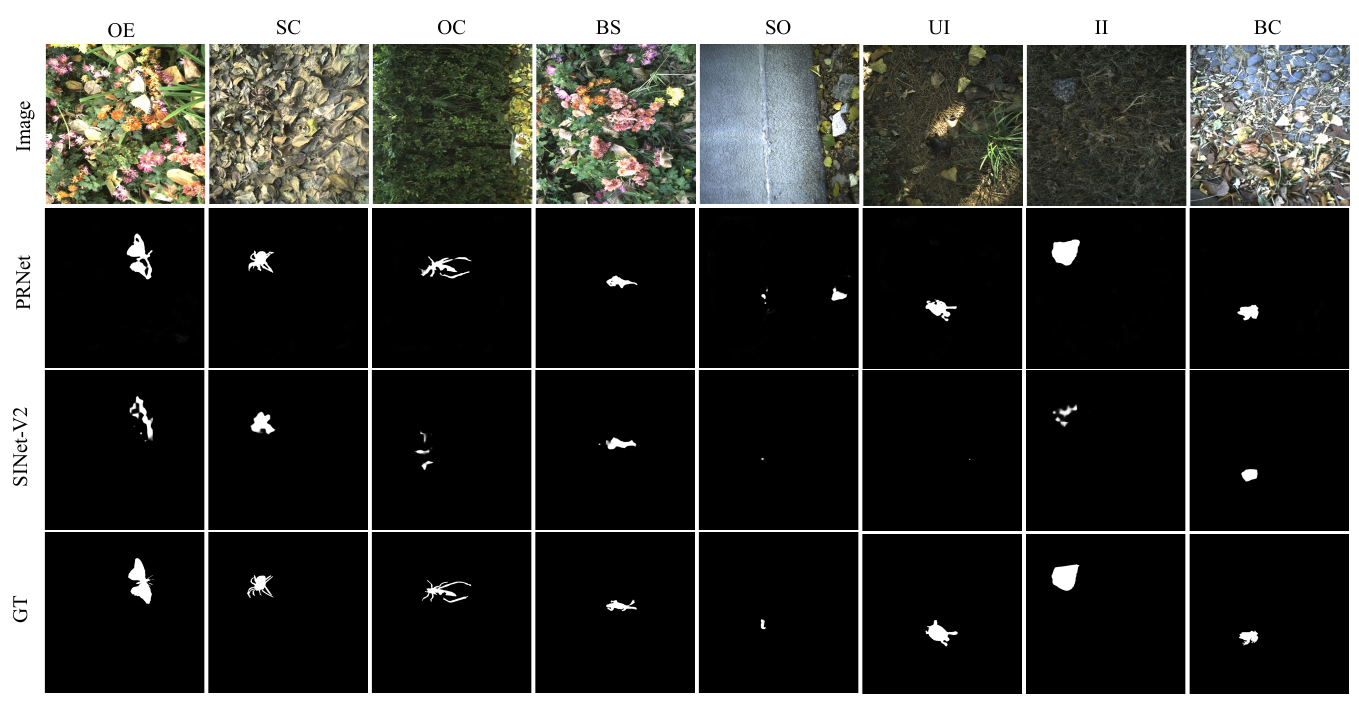}
  \vspace{-4mm}
  \caption{Visualization of detection results under various challenging scenarios.}
  \label{fig:s3}
\end{figure*}

\subsection{Evaluation Results}
\noindent
\textbf{Overall Performance.}  
Table~\ref{tab:benchmark} summarizes the quantitative results. Notably, PRNet achieves the best overall performance in terms of \(E_\xi\) (0.926), \(F_\beta\) (0.698), and MAE (0.002). IdeNet also delivers competitive results; both models are based on Vision Transformer architectures, highlighting their effectiveness for COD tasks. Interestingly, although C2FNet-V2 is not Transformer-based, it performs remarkably well and even surpasses IdeNet in \(E_\xi\), \(S_\alpha\), and \(F_\beta\). A particularly notable observation is that PCNet achieves the highest score on \(S_\alpha\) (0.855), yet falls significantly behind Transformer-based methods in the remaining metrics. Figure~\ref{fig:s3} presents sample prediction results under selected challenging attributes. The figure illustrates that the diverse and challenging attributes in the MCOD dataset pose significant difficulties for existing camouflaged object detection methods, highlighting the advantages of MCOD in advancing this research field.

\begin{table}[!t]
\centering
\caption{Comparison of results under RGB and MSI inputs.}
\label{tab:RGB-MSI}
\begin{tabular}{
    >{\centering\arraybackslash}m{1.8cm} | 
    >{\centering\arraybackslash}m{1.0cm} | 
    *{4}{>{\centering\arraybackslash}m{0.8cm}} 
}
\toprule
\textbf{Method} & \textbf{Input} & \boldmath$E_{\xi}\uparrow$ & \boldmath$S_{\alpha}\uparrow$ & \boldmath$F_{\beta}\uparrow$ & \boldmath$M\downarrow$ \\
\midrule

\multirow{2}{*}{\centering SINet\cite{r11}} 
 & RGB & 0.695 & 0.601 & 0.335 & \textbf{0.005} \\
 & MSI & \textbf{0.758} & \textbf{0.616} & \textbf{0.369} & 0.006 \\
\cmidrule(lr){1-6} 

\multirow{2}{*}{\centering CODCEF\cite{r34}} 
 & RGB & 0.718 & 0.632 & 0.359 & 0.005 \\
 & MSI & \textbf{0.763} & \textbf{0.677} & \textbf{0.444} & \textbf{0.004} \\
\cmidrule(lr){1-6} 

\multirow{2}{*}{\centering C2FNet-V2\cite{r33}} 
 & RGB & 0.885 & 0.743 & 0.553 & 0.009 \\
 & MSI & \textbf{0.913} & \textbf{0.810} & \textbf{0.654} & \textbf{0.008} \\
\cmidrule(lr){1-6}

\multirow{2}{*}{\centering PCNet\cite{r32}} 
 & RGB & 0.397 & 0.788 & 0.149 & 0.005 \\
 & MSI & \textbf{0.633} & \textbf{0.855} & \textbf{0.386} & \textbf{0.003} \\
\bottomrule
\end{tabular}
\vspace{-0.5\baselineskip}
\end{table}

\noindent
\textbf{Qualitative Analysis of Multispectral Modality.} A central question in multispectral camouflaged object detection is whether the multispectral modality can offer performance advantages over RGB-based models. To investigate this, we conducted a series of comparative experiments to assess the impact of RGB and multispectral data on the COD task. Specifically, we trained and tested CODCEF, C2FNet-V2, and PCNet using both RGB and multispectral images. The results, summarized in Table~\ref{tab:RGB-MSI}, demonstrate consistent performance improvements with multispectral input. For example, C2FNet-V2 achieved an \(S_\alpha\) of 0.743 and an \(F_\beta\) of 0.553 when trained on RGB images. With multispectral data, these metrics increased to 0.810 (a 9.05\% improvement) and 0.654 (an 18.26\% improvement), respectively. We further analyzed the performance of CODCEF and C2FNet-V2 across the eight challenging attributes before and after incorporating the multispectral modality, as shown in Table~\ref{tab:challenge_improvement}. Both models exhibited notable improvements under all attributes. For instance, C2FNet-V2 showed a 13.4\% increase in \(S_\alpha\) for the ``Small Object'' attribute, rising from 0.647 to 0.734. Similar performance gains were observed across the remaining seven attributes. These results underscore the effectiveness of multispectral features in enhancing camouflaged object detection, aligning with the spectral response curve analysis in Fig.~\ref{fig:MSI}.

\begin{table}[!t]
\centering
\caption{The\textbf{ \(S_\alpha\)} scores across the 8 challenging attributes under RGB and MSI inputs.}
\label{tab:challenge_improvement}
\setlength{\tabcolsep}{3pt} 
\begin{tabular}{
    >{\centering\arraybackslash}m{1.8cm} | 
    >{\centering\arraybackslash}m{0.8cm} | 
    *{8}{>{\centering\arraybackslash}m{0.45cm}} 
}
\toprule
\textbf{Method} & \textbf{Input} & 
\textbf{SO} & \textbf{BS} & \textbf{CB} & \textbf{OE} & 
\textbf{II} & \textbf{UI} & \textbf{OC} & \textbf{SC} \\
\midrule

\multirow{2}{*}{C2FNet-V2\cite{r33}} 
 & RGB & .647 & .728 & .726 & .634 & .810 & .731 & .749 & .790 \\ 
 & MSI & \textbf{.734} & \textbf{.774} & \textbf{.798} & \textbf{.685} & \textbf{.837} & \textbf{.768} & \textbf{.813} & \textbf{.826} \\
\addlinespace[0.1cm]
\cmidrule(lr){1-10}

\multirow{2}{*}{CODCEF\cite{r34}} 
 & RGB & .555 & .619 & .585 & .558 & .736 & \textbf{.661} & .637 & .670 \\
 & MSI & \textbf{.597} & \textbf{.674} & \textbf{.717} & \textbf{.624} & \textbf{.737} & .651 & \textbf{.687} & \textbf{.722} \\
\bottomrule
\end{tabular}
\vspace{-0.5\baselineskip}
\end{table}

\begin{figure}
  \includegraphics[width=1\linewidth]{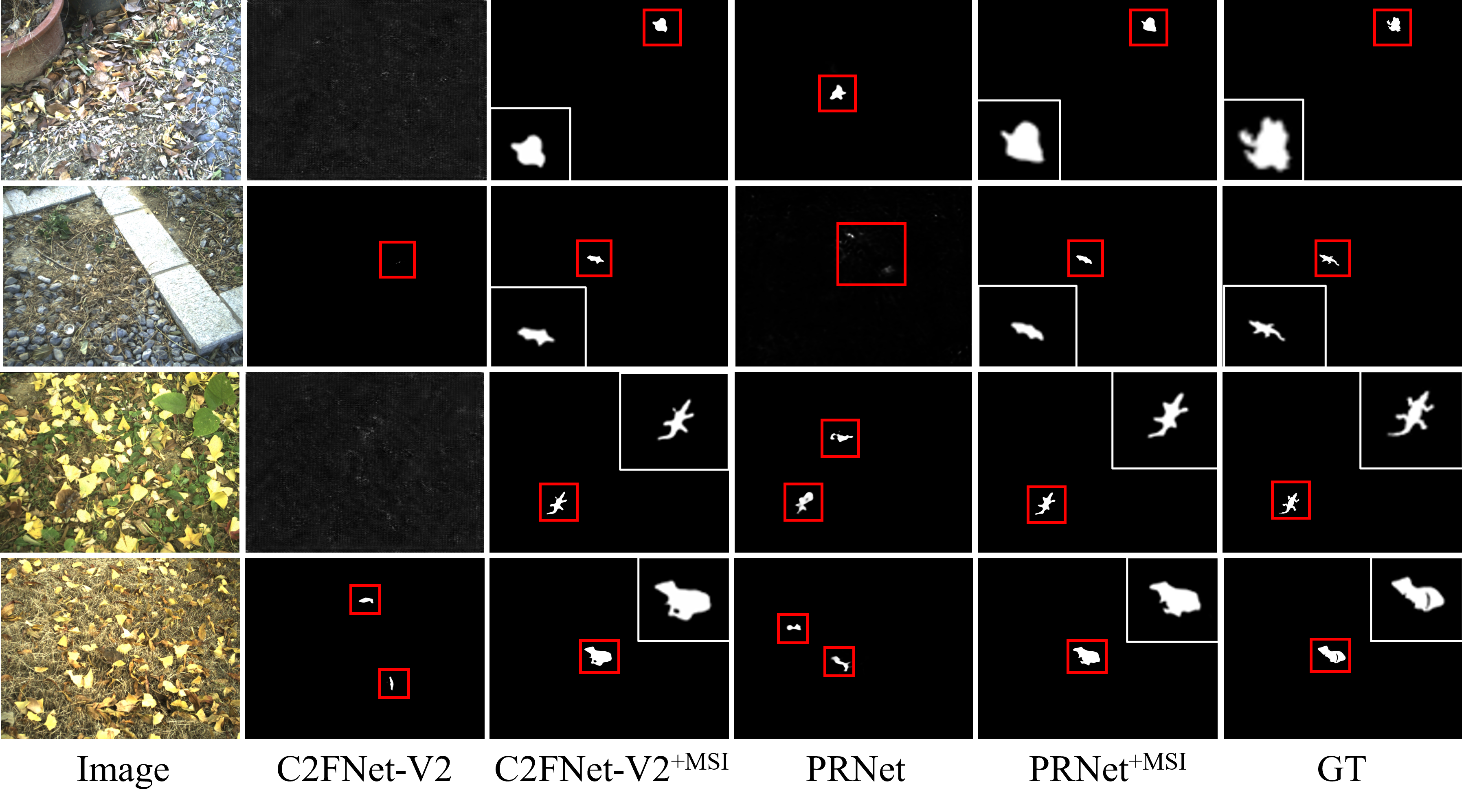}
  \vspace{-6mm}
  \caption{Improvements in missed and false detections enabled by multispectral modality.}
  \label{fig:qualitative_comparison}
\end{figure}

\begin{figure}[t]
  \includegraphics[width=1\linewidth]{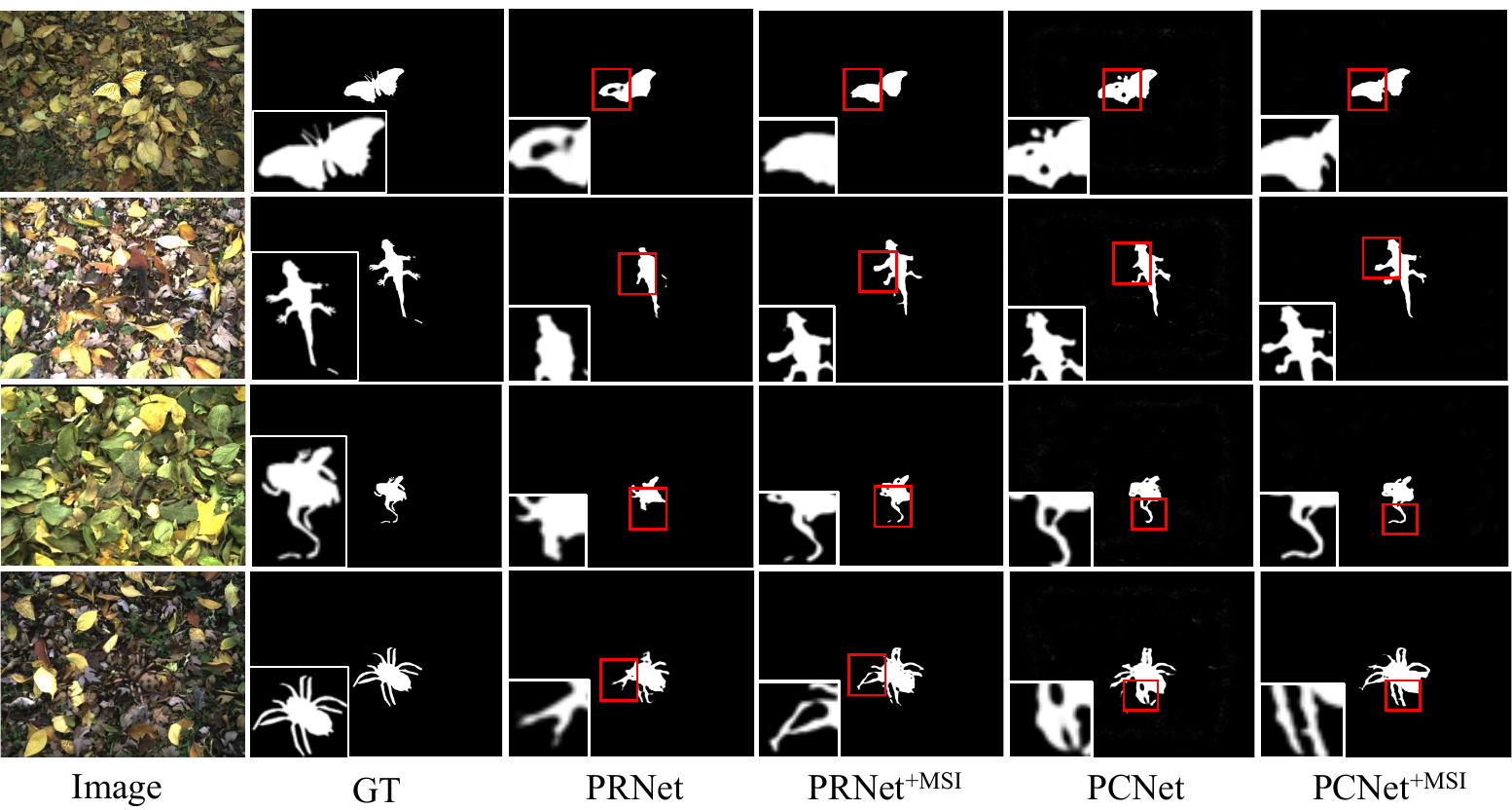}
  \caption{Improved delineation of object boundaries and contours enabled by the multispectral modality.}
  \label{fig:f5}
\end{figure}

\begin{figure}
  \includegraphics[width=1\linewidth]{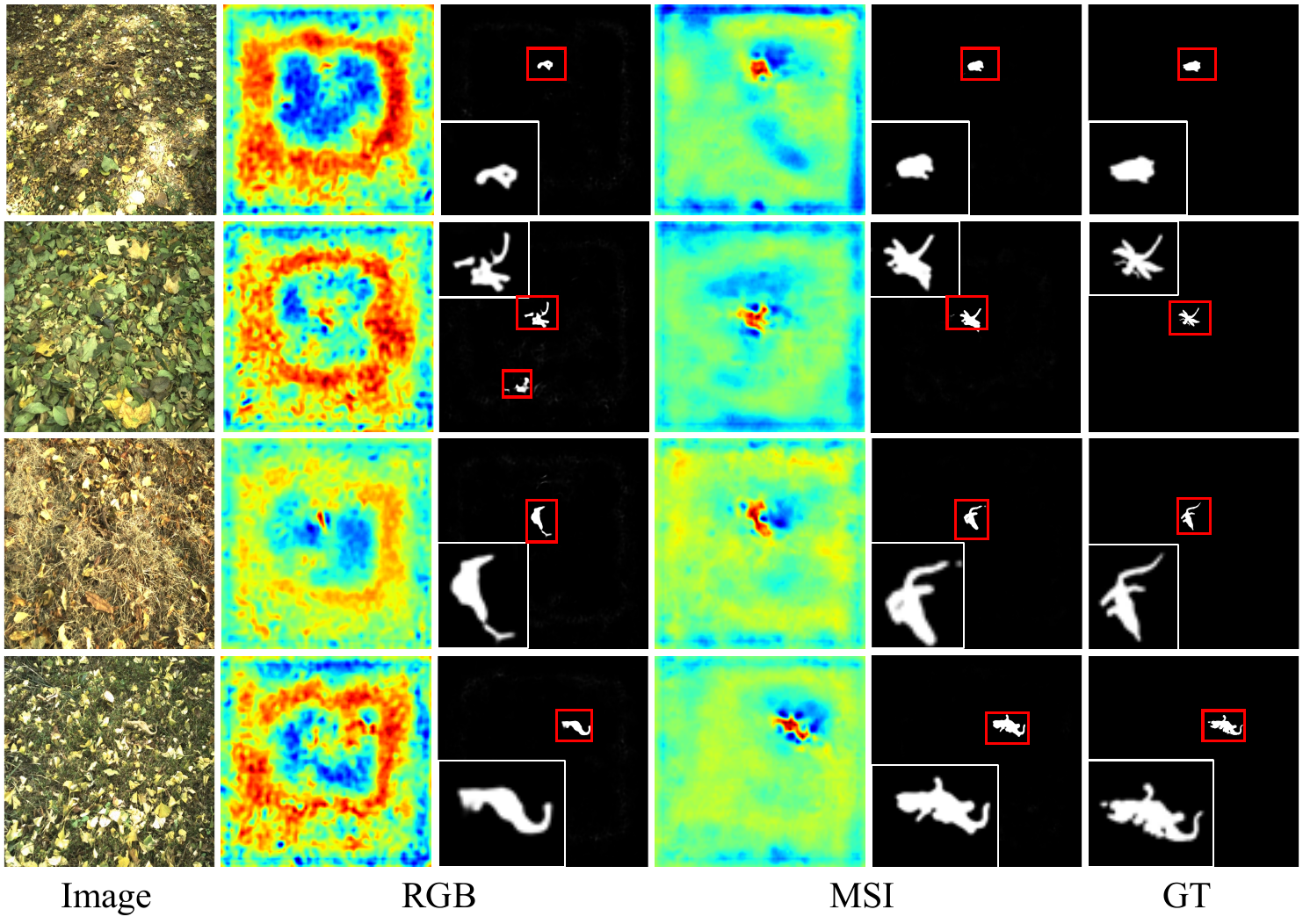}
  \vspace{-6mm}
  \caption{Visualization of intermediate feature maps under RGB and MSI inputs.}
  \label{fig:spectral_features}
  \vspace{-4mm}
\end{figure}

\noindent
\textbf{Qualitative Comparison of RGB and MSI Inputs.} 
Building upon the quantitative analysis, we present qualitative evaluation results on MCOD. To intuitively assess the impact of multispectral information on camouflaged object detection, Fig.~\ref{fig:qualitative_comparison} shows the prediction outputs of two advanced models—PRNet and C2FNet-V2—using both RGB and multispectral images as input. As illustrated, RGB-based models often fail to detect or accurately delineate objects under challenging conditions. In contrast, models incorporating multispectral inputs exhibit enhanced object localization and sharper contour delineation. This improvement is attributed to the rich spatial–spectral information inherent in multispectral data, which enables the models to learn more discriminative features. As shown in Fig.~\ref{fig:f5}, multispectral input not only significantly reduces false positives and missed detections but also enhances the accuracy of boundary prediction. Furthermore, Fig.~\ref{fig:spectral_features} demonstrates that multispectral cues substantially improve the models’ ability to identify camouflaged objects in complex backgrounds and refine object boundaries, thereby increasing overall prediction accuracy.

\noindent
\textbf{Attribute-Based Evaluation.} 
To further analyze the performance of various camouflaged object detection models, we conduct attribute-based evaluations on the eight challenge attributes defined in the MCOD dataset. The results show that current COD methods still maintain relatively good performance under challenges such as shape complexity and occlusion. However, their performance deteriorates significantly when faced with small objects, extreme illumination conditions, background similarity, and complex backgrounds. These findings suggest that such scenarios remain the most difficult in camouflaged object detection and highlight the need for further advancements in existing approaches. The results reveal that existing COD methods experience notable performance degradation in scenarios involving small objects, extreme illumination, background similarity, and complex backgrounds. These challenges remain the most difficult in COD tasks and highlight the need for continued methodological advancements. By assessing the performance of existing methods across diverse challenging attributes, we gain clearer insight into the complexities inherent in camouflaged object detection. These findings offer valuable guidance for shaping future research directions in this field.

\noindent
\textbf{Discussion.} 
The evaluation results point to several key research directions: \textbf{(1) Feature extraction networks.} Most top-performing models in Table~\ref{tab:benchmark} are based on Vision Transformer architectures, suggesting that the development of more effective feature extraction networks is critical for advancing COD performance. However, despite employing powerful backbones, these models still fall short of achieving satisfactory results. This may be attributed to the fact that their architectures are primarily designed for RGB images, without accounting for the unique feature extraction requirements of other modalities. \textbf{(2) Multispectral information.} Multispectral data offer rich spatial–spectral cues that are highly beneficial for COD tasks. Nonetheless, this information has often been underutilized, overshadowed by the recent dominance of Vision Transformers and the trend toward more complex network architectures with increased computational demands. Our experimental results, however, demonstrate that incorporating multispectral modalities can significantly improve detection performance without increasing model complexity. These findings underscore the urgent need to integrate multispectral features into general COD models.

\section{Conclusion}
In this paper, we present MCOD, the first large-scale benchmark for multispectral camouflaged object detection. MCOD comprises 1,527 carefully registered and calibrated multispectral images collected from diverse real-world scenarios. Each image is accompanied by pixel-level ground-truth annotations and eight challenging attribute labels. We conduct a comprehensive evaluation of state-of-the-art camouflaged object detection models and demonstrate that the multispectral modality offers significant advantages over traditional RGB approaches. We anticipate that the MCOD benchmark will serve as a solid foundation for future research in multispectral camouflaged object detection.


\bibliographystyle{ACM-Reference-Format}
\bibliography{MCOD}

\appendix

\end{document}